\pdfoutput=1

\documentclass[11pt]{article}

\usepackage[]{emnlp2021}

\usepackage{times}
\usepackage{latexsym}

\usepackage[T1]{fontenc}

\usepackage[utf8]{inputenc}

\usepackage{microtype}

\usepackage{times}
\usepackage{placeins}
\usepackage{amsmath}
\usepackage{booktabs}
\usepackage{latexsym}
\usepackage{float}
\usepackage[linesnumbered,lined]{algorithm2e}
\usepackage[T1]{fontenc}
\usepackage[utf8]{inputenc}

\usepackage{microtype}

\usepackage{graphicx}
\usepackage{amsmath}
\usepackage{enumitem}
\usepackage{subfig}
\usepackage[normalem]{ulem} 
\usepackage{makecell}
\usepackage{array}
\usepackage{multirow}
\usepackage[symbol]{footmisc}

\title{On Text Style Transfer via Style Masked Language Models}

\author{Sharan Narasimhan, Pooja Shekar,  Suvodip Dey, Maunendra Sankar Desarkar \\
Department of Computer Science and Engineering \\ 
Indian Institute of Technology Hyderabad, India\\\\
\texttt{\{sharan.n21, poojashekar15\}@gmail.com}\\
\texttt{\{cs19resch01003, maunendra\}@iith.ac.in}}

\begin{document}
\maketitle

\begin{abstract}
    Text Style Transfer (TST) is performable through approaches such as latent space disentanglement, cycle-consistency losses, prototype editing etc. The prototype editing approach, which is known to be quite successful in TST, involves two key phases a) Masking of source style-associated tokens and b) Reconstruction of this source-style masked sentence conditioned with the target style.
    We follow a similar transduction method, in which we transpose the more difficult direct source to target TST task to a simpler Style-Masked Language Model (SMLM) Task, wherein, similar to BERT \cite{bert}, the goal of our model is now to reconstruct the source sentence from its style-masked version. 
    We arrive at the SMLM mechanism naturally by formulating prototype editing/ transduction methods in a probabilistic framework, where TST resolves into estimating a hypothetical parallel dataset from a partially observed parallel dataset, wherein each domain is assumed to have a common latent style-masked prior. To generate this style-masked prior, we use ``Explainable Attention'' as our choice of attribution for a more precise style-masking step and also introduce a cost-effective and accurate ``Attribution-Surplus'' method of determining the position of masks from any arbitrary attribution model in O(1) time.
    We empirically show that this non-generational approach well suites the ``content preserving'' criteria for a task like TST, even for a complex style like Discourse Manipulation. Our model, the Style MLM, outperforms strong TST baselines and is on par with state-of-the-art TST models, which use complex architectures and orders of more parameters.
\end{abstract}

\section{Introduction}
Text Style Transfer (TST) can be thought of as a subset of the Controllable Language Generation Task \cite{hu} with tighter criteria. TST involves converting input text possessing some source style into an output possessing the target style, meanwhile preserving style-independent content and maintaining fluency. Style is usually defined by the class labels present in an annotated dataset, commonly considered to be Sentiment, Formality, Toxicity etc. We consider the unsupervised setting where only non-parallel datasets are available, i.e. the style-transferred output is not available. Past work focuses on various common paradigms such as disentanglement \cite{hu, crossaligned}, cycle-consistency losses \cite{yi, luo, cyclemulti, direct}, induction \cite{epaae, daae} etc. We focus on a sentence to sentence ``transduction'' (or prototype editing) method, a solution which naturally emerges when following a probabilistic formulation consisting of a single transduction model with a latent prior consisting of a style-absent corpus. Our approach is also inspired by BERT's MLM \cite{bert}\footnote[1]{A key difference to BERT, ours is not using a self-supervised approach of pre-training over large corpora and fine-tuning for a downstream task.}, modified to incorporate style information to enable TST. This notion can be seen from various viewpoints.
\begin{table}
\small
\centering
\begin{tabular}{p{0.8cm}p{2.5cm}p{2.5cm}} 
\hline
\textbf{Task}                                    & \begin{tabular}[c]{@{}c@{}}\textbf{Positive }\\\textbf{to Negative}\end{tabular} & \begin{tabular}[c]{@{}c@{}}\textbf{Contradiction}\\\textbf{~to Entailment}\end{tabular}                 \\ 
\hline
\textbf{Input}                                                 & this movie is by far one of the best urban crime dramas i 've seen .             & a woman is sitting outside at a table using a knife to cut into a sandwich . a woman is sitting inside  \\
\begin{tabular}[c]{@{}c@{}}\textbf{Style}\\\textbf{Masked}\end{tabular} & this movie is by mask one of the mask urban crime mask i 've seen .              & a woman is sitting outside at a table using a knife to cut into a sandwich . a woman mask mask mask     \\
\textbf{Output}                                                & this movie is by far one of the worst urban crime garbage i 've seen .           & a woman is sitting outside at a table using a knife to cut into a sandwich . a woman is a outside       \\
\hline
\end{tabular}
\caption{Examples of Sentiment and Discourse TST by the SMLM on the IMDb and SNLI datasets respectively.}
\vspace{-0.2in}
\end{table}

\textbf{Another MLM.} 
By making use of various text attribution models, it is possible to determine the fraction of tokens that significant contribution to the style label. Masking out these tokens and training a bi-directional transformer encoder with self-attention with the task of reconstructing the original source style sentence is intuitive and closely resembles BERT-based MLM's \cite{bert, roberta}. Instead of random masking or perplexity based masking \cite{roberta}, by incorporating style-masking and concatenation of control codes that inform the encoder about the target style needed, the resultant MLM now resembles a rough TST model. This notion of "filling in only the style words" also does well to automatically ensure that other content words are preserved, and fluency of the original sentence is maintained. Unlike most generational approaches, this method has an easier objective by inferring what to fill in by looking at both directions, similar to BERT.

\textbf{Parallel vs Non Parallel.} A "parallel" dataset, in which every instance in the dataset has associated sentences with all given style labels, is naturally much easier to salvage but harder to come by. In this case, TST resolves into the supervised task of simply learning the input distribution corresponding to this parallel dataset. In reality, most available datasets are non-parallel/unsupervised (E.g. Yelp). TST models must infer the notion of "Style" present in the dataset using non-aligned sentences of the same styles, differing in content, with no direct supervision signal. \\
\textbf{A Hypothetical Parallel.}
In this case, TST resolves into the task of estimating the missing portion of a hypothetical parallel dataset without explicit knowledge of the complete input distribution as in the parallel case. We adopt a probabilistic formulation of TST, in which we assume each domain in this hypothetical parallel dataset originates from a common discrete latent prior, a latent style-masked prior in the case of SMLM. Salvaging a non-parallel dataset to learn a simple "Same-Style Reconstruction" task using the SMLM also indirectly approximates the cross-domain TST task, which we only need to fine-tune for a single epoch to make it comparable to larger SOTA models. 
\\


\textbf{Contributions.}
a) We introduce the Style Masked Langauge Model (SMLM), another type of MLM capable of being of strong TST model by initially pretrained on an unsupervised same-style reconstruction task and fine-tuned for TST metrics.
b) We use an accurate and effective style masking step and present an analysis to support our choice.
c) We empirically show that the SMLM beats strong baselines and compares to state of the art, using far lesser parameters.
d) We adopt a novel "Discourse" TST task by salvaging NLI datasets and show that our style-masking is also capable of working with more complex "Inter-sentence' styles to manipulate the flow of logic.

\section{Related Work}
\citet{survey1} and \citet{survey2} provide surveys detailing the current state of TST and lay down useful taxonomies to structure the field. In this section, we only discuss recent work similar to ours (a prototype editing approach) assuming the same unsupervised setting.

\citet{simple} present the earliest known work using the prototype editing method, in which a "delete" operation is performed on tokens based on simple count-based methods, and the retrieval of the target word is done by considering  TF-IDF weighted word overlap. \citet{padded} first train MLMs for the source and target domains and perform TST by first masking text spans where the models disagree (in terms of perplexity) the most, and use the target domain MLM to fill these spans. \citet{maskinfill} introduce the Attribute-Conditional MLM, which most closely aligns to the working of the MLM, also uses an attention classifier for attribution scores, a count and frequency-based method to perform masking, and a pretrained BERT model fine-tuned on the TST task. \citet{stable} and \citet{masktransformer} also follows roughly the same pipeline but uses a generational transformer encoder-decoder approach and also fine-tunes using signals from a pretrained classifier. \citet{point} uses a hierarchical reinforced sequence operation method is used to iteratively revise the words of original sentences. \citet{tag} uses n-gram TF-IDF based methods to identify style tokens and modify them as either "add" or "replace" TAG tokens, which are then substituted by the decoder to perform TST. Similar to the SMLM, \cite{cycle_rl} also uses attribution-based methods from a self-attention classifier. However, they use an LSTM \cite{lstm} based approach, one to generate sentences from each domain.

\section{Method}
\subsection{Notation}
Let $S$ denote the set of all style labels for a supervised dataset  $D$ of the form $\{(x_0,l_0),(x_1,l_1)\dots(x_n,l_n)\}$ where $x_i$ denotes the input sentence and $l_i \in S$ denotes the label corresponding to $x_i$. The set of all sentences of style $s$ in $D$ is denoted by $\hat{x^s} = \{x_j : \forall{j}\ where (x_j,s)\in{D} \}$. We use a special meta label $m_s$ to represents the "style-masked" label class having $s$ the original style. Subsequently, $x^{m_s}$ refers to the set of all style-masked sentences which had source style $s$. The set of all style-masked sentences from $D$ is given by $x^m=Union(x^{m_s}:\forall{s}\in{S})$.  Let $x^s$ denote the model's estimation of $\hat{x^s}$.

\begin{figure}[t]
\includegraphics[width=8cm]{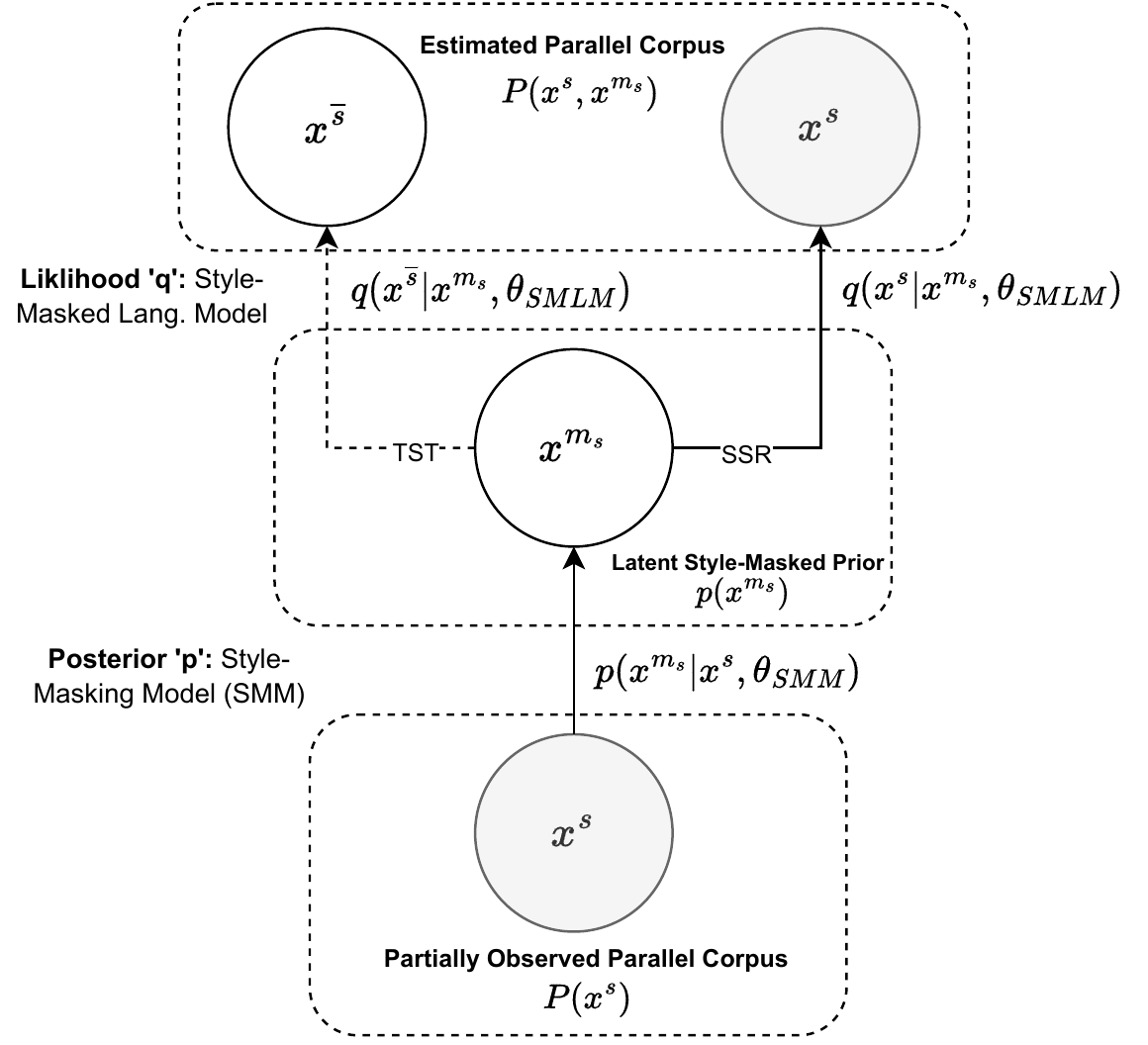}
\caption{Probabilistic overview of Transduction Model assuming common Latent Style-Masked Variable as prior.}
\label{fig:prob-overview}
\end{figure}

\textbf{Probabilistic Overview.} We treat the non-parallel dataset as a partially observed parallel dataset, where the SMLM has to estimate the entire joint distribution $P(x^s, x^{\hat{s}})$. The SMLM assumes that every sentence from each style domain is a result of, \textbf{a)} sampling from a latent style-masked prior, $p(x^{m_s})$, which we estimate using the Style-Masking model, $p(x^{m_s}|x^s, \theta_{SMM})$ (conditional posterior), and is then \textbf{b)} conditionally reconstructed to form the sentence with target style using the SMLM, $q(s^s, x^{\overline{s}}|x^{m_s})$ (the style-conditioned likelihood). The posterior in this case is easily estimatable by using various intuitive methods. The likelihood function, represented by the SMLM, has a simple task of performing the same-style reconstruction task using a non-parallel dataset. Using control codes to guide which domain to sample from, enables the SMLM to also estimate the unseen section of the hypothetical parallel and therefore work as a rought TST model. The probabilistic formulation of the overall model is given in Fig. \ref{fig:prob-overview}.

\subsection{Explainable Attention as Attribution Markers}
Many prototype editing methods use Vanilla Attention (VA) as attribute scores \cite{maskinfill, memories, masktransformer}. However, it has been shown that attention is not explanation, at least for the purpose of human-aligned interpretability \cite{jain}. VA does not correlate well with other well-known attribution methods (such as Integrated Gradients \citet{ig}). We instead use "Explainable Attention" (EA) scores from a Diversity-LSTM classifier \cite{mohankumar, nema2} which have been shown to correlate better with other attribution methods. EA scores as attribution also correlates better with human judgement as compared to VA. We discuss the intuitive reasoning as to why VA over LSTM hidden states fail as attribution scores, the intuition of the Diversity-LSTM, its mechanism and loss equations in Section \ref{sec:ea}. We also quantitatively compare the efficacy of the style-masking step between EA and VA in Table \ref{tab:attri-compare}.

\subsection{The Style-Masking module}
Even with having accurate attribution scores, effective style-masking requires careful selection of a policy which satisfies certain criteria. The primary criteria being that only tokens which significantly contribute to the style of a sentence must be masked, and leaving other tokens, to ensure content is also preserved. Similar to the masking policy in \citet{masktransformer}, it is natural to consider a "top $k$ tokens" scheme in which the top $k$ tokens with highest attribution are masked. However, this static approach fails for sentences which do not have $k$ style-contributing tokens, leading to either partial style masking or erroneous masking of content tokens. For the same reason, even a sentence length aware scheme such as "top 15\%" masking fails. Furthermore, all such policies require sorting, leading to O(nlogn) time complexity for style masking of each sentence in a batch.
\\
\textbf{An "Attention Surplus" policy. }
Let $A=\{A_i\dots A_n\}$ denote the attention distribution of a sentence of size $n$. Intuitively, we can reason that all "special" tokens which might contribute more to style should have an attribution greater than the average base attribution of the sentence, given by $A^{mean} = 1/n$. Generalising this further, We refer to tokens with $A^i \geq A^{baseline}$ as tokens with "attention surplus" w.r.t to a sentence-length sensitive baseline attention $A^{baseline}$ given by:

\begin{align}
    A^{baseline}=(1+\lambda_{\epsilon})*A^{mean}
\end{align}

where $\lambda_{\epsilon}$ is a hyperparameter of range $0-1.0$. This chosen threshold $A^{baseline}$ is sensitive to the sentence length as well and subsequently ensures that the number of style-significant tokens can be dynamically determined, without need of an elaborate algorithm. As a sanity check, we observe that even in the adversarial case where all tokens might be equally important to style, $A$ resolves into a $UniformDistribution(n)$ and our policy correctly resorts to masking all tokens\footnote{Assuming $\lambda_{\epsilon}=0$, whereas in practice we find $\lambda_{\epsilon}=0.15,0.5$ giving optimal masking for the Sentiment and Discourse TST, respectively. More is discussed in \ref{sec:eps-effect}}. Let $Mask$ denote the token mask matrix of size $n$ initialised with zeros.
\begin{align}
    \text{Mask}[A_i \geq A^{baseline}] = 1
\end{align}
Using a vectorised approach as given in the above equation, we can achieve style-masking for an entire batch in O(1) as compared to O(nlogn) per sentence in previous approaches.

\subsection{A Style-aware MLM via Transformer Encoder Block}
\textbf{The SMLM.} For the sake of convenience in notation, we assume binary style labels, $S=\{0,1\}$\footnote{In theory, this can be extended to any number of styles.}. The task of TST now resolves into estimating $q(\hat{x^{\overline{s}}}|x^{m_s}, \theta_{SMLM})$. The style-masking module from the previous step gives us access to the discrete prior $p(x^{m_s}$) i.e. available as pairs of the form $(x^s_i, x^{m_s}_i)$. The task of reconstructing a sentence $x^{m_0}_i$ e.g. "The food was <blank>" into the original sentence $x^0_i$, e.g. "The food was good" resembles a Masked Language Model (MLM) objective similar to \citet{devlin} where additionally the target style is contextualised as well. We thus refer to this task of "Style-aware" MLM (SMLM), which consist of estimating two distributions i.e. $q(x^{\overline{s}}|x^{m_s}, \theta_{SMLM})$, and $q(x^{s}|x^{m_s},\theta_{SMLM})$. 
\\
\textbf{Control Codes.} We concatenate two special meta-tokens $[dst]$, $[src]$ to $x^{m_s}_i$ i.e. the source and destination style labels respectively. This is denoted by the function $concat(x^{m_s}_i,[src],[dst])$. Choice of $[src]$, $[dst]$ triggers the SMLM model to sample different distributions, either $q(x^{s}|x^{m_s})$ or $q(x^{\overline{s}}|x^{m_s}$), corresponding to the "Same-Style" Reconstruction and TST Task respectively. The final model is defined as:
\begin{align}
    \nonumber q_{SMLM}= 
\begin{cases}
    q(x^{s}_i|x^{m_s}_i,\theta_{SMLM}),& \text{if } [src]=[dst] \\
    q({x^{\overline{s}_i|}x^{m_s}_{i}},\theta_{SMLM}),& \text{if } [dst] = \overline{[src]}
\end{cases}
\end{align}

\textbf{2-Stage Training.} Similar to \citet{direct}, we use a two stage method of training our TST model. 
In the first stage, we bootstrap a rough TST model and then finetune to improve selected metrics for the TST task.
\subsubsection{Bootstrapping on Same Style Reconstruction Task}
 The SMLM model is first trained to estimate the likelihood $q(x^{s}|x^{m_s})$ and therefore set $[src]=[dst]=s$ in $concat$. This corresponds to the task of reconstructing original sentences $x^s$ from their corresponding style-masked sentence $x^{m_s}$ produced by the SM module. Due to the nature of the SMLM task in which input and output pairs ($x^{m_s}$, $x^{s}$) always have the same length, the Transformer \cite{vaswani} encoder block posses as a natural choice since it enforces same-dimension output transformations \footnote{This however constrains the output size to always match the input. Consequently, our SMLM model is not generational. In section \ref{sec:ablation}, we further motivate this decision.}. The Transformer Encoder Block $E$ minimizes the NLL given below:
 \begin{align}
    \nonumber L_{BS}(\theta_{SMLM}) =  - \mbox{log}  q_{SMLM} (\hat{x^s} | x^s)
\end{align}

\subsubsection{Fine-tuning for TST}
We now fine-tune the SMLM's performance for the certains metric in the TST task i.e. $q(x^{\overline{s}}|x^{m_s}$) which is triggered by setting $[dst]=\overline{[src]}$. We fine-tune for TST\% by jointly training a style classifier $cls$ by minimising the NLL taking the
same-style reconstructions of SMLM i.e. $x^s$ as input.\footnote{We use a single feed-forward layer with input being the average of the last layer embeddings of $X^s_i$ (excluding the meta-labels).}
 \begin{align}
    \nonumber L_{cls}(\theta_{cls}) =  - \mbox{log} p_{cls} (s| x^{s})
\end{align}
This classifier now provides a supervision signal that the SMLM uses to align $x^{\overline{s}}$ to style attribute $\overline{s}$. Whereas the discriminator tries to discourage this. This is formulated as min max objective $L_{FT}$ of the form:
 \begin{align}
    \nonumber \min_{\theta_{SMLM}} \max_{\theta_{cls}} -log  p_{cls}(\overline{s}|x^{\overline{s}})
\end{align}
We do not fine-tune for content-preservation or fluency as we do not observe any lacking in these metrics after bootstrapping.

\begin{figure}[t]
\includegraphics[width=8cm]{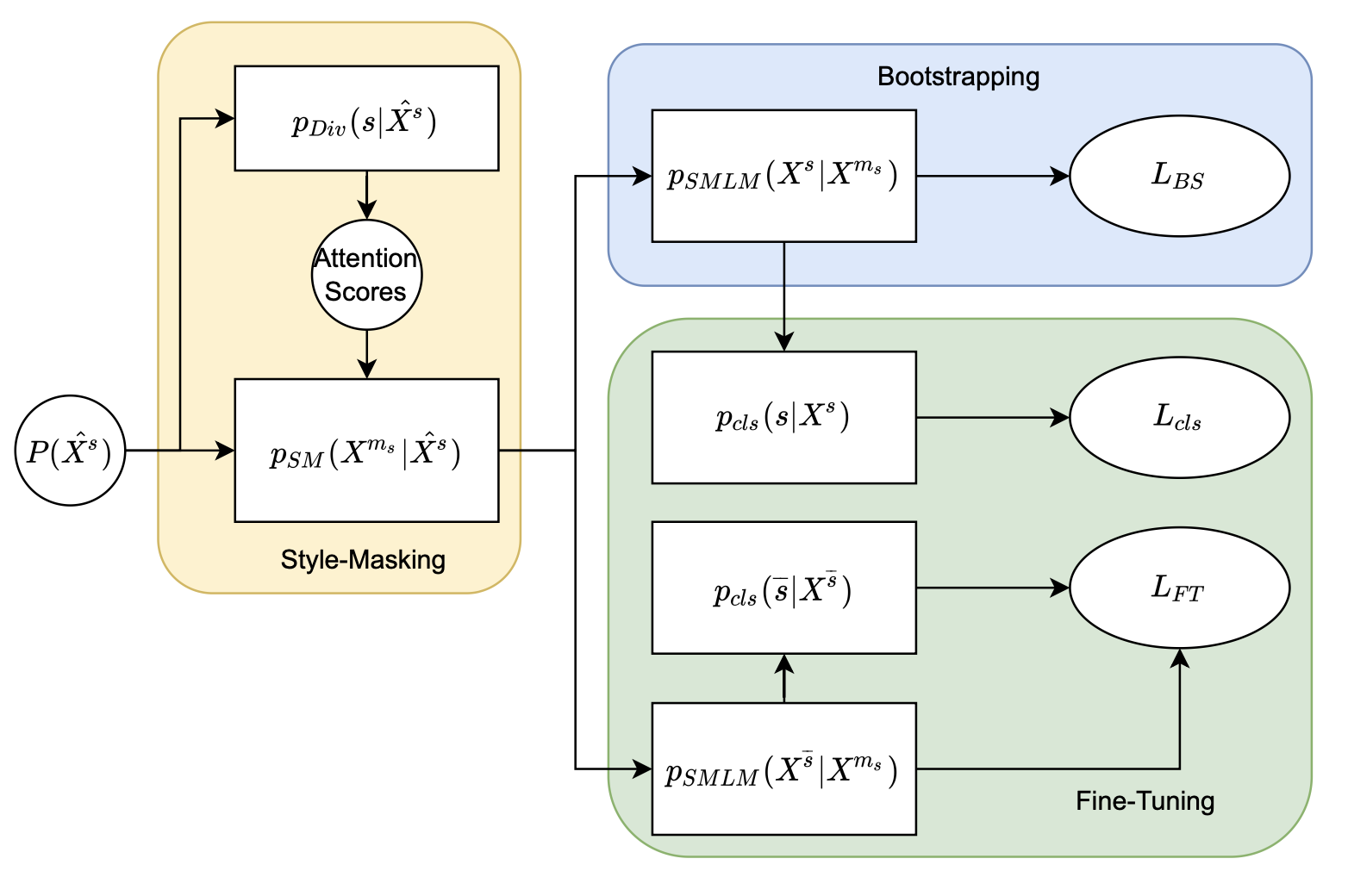}
\caption{Overview of Model Architecture considering sentences of style $s$ i.e. $x^s$. In reality, this is applied all styles $x^s, \forall{s}\in{S}$}
\vspace{-0.2in}
\end{figure}

\section{Datasets and Tasks}
We report the split and label wise statistics of each dataset in \ref{dataset_details}.\\
\textbf{Sentiment TST: }Following many past studies, we evaluate our model for the sentiment TST task using three review datasets, Yelp, Amazon and IMDb  All three datasets are annotated with two labels corresponding to positive or negative reviews and are non-parallel. \footnote{For Yelp, Amazon and IMDb, we used the pre-processed version specified in https://github.com/yixinL7/Direct-Style-Transfer.}

\textbf{Discourse TST: }To illustrate that our model also performs well in such situations, we introduce the Discourse TST task by performing TST on the SNLI \cite{snli} dataset. Each instance consists of two sentences, which either contradict, entail (or agree) or are neutral (no relationship) to each other. We consider the task of converting the discourse style between two sentences, from contradiction to entailment and vice-versa. Some TST tasks are more complex than others and have different levels of granularity \cite{styleptb}. Unlike the sentiment task, which is "intra-sentence", where the style can be attributed to a select set of words, the discourse task is "inter-sentence" and requires the model to be cognizant of the context and detect the flow of logic rather than just a few style words (especially for the Contradiction to Entailment Task).

\section{Analysis of Style-Masking approach}
In this section, we evaluate and justify our choice of style-masking architecture, i.e. "Explainable Attention" + "Attention-Surplus" masking policy. We explore the question \textit{"What actually constitutes a good style-masking step?"}. Intuitively, we can reason that our style-masking approach must a) produce accurate attribution scores for each token b) use an appropriate masking policy that determines which tokens to mask using these attribution scores. The final style-masked sequences (latent prior to the SMLM) must actually be devoid of style information and accurate, i.e. not done at the expense of content information.

\subsection{Quant. Analysis of Various Attribution Methods}
We consider various other attribution methods for our analysis i.e. Vanilla Gradients, Integrated Gradients \cite{ig}, Vanilla Attention, Attention * X (or inputs) and Explainable Attention \cite{mohankumar}. We do not consider techniques such as LIME \cite{lime}, LRP \cite{lrp}, DeepLIFT \cite{deeplift} as they are relatively more computationally expensive during inference time. As the style-masking policy, use the "attribution-surplus" policy to determine the final style-masked sequences.
\\
\textbf{In Table \ref{tab:attri-compare}, } we compute the Accuracy\% and s-BLEU on the final style-masked sequences produced by each attribution method on the test split for all four datasets. We can reason that an ideal style-masking method should be able to produce sentences that completely mask out style, thereby fooling a pretrained classifier (minimizing its Accuracy\%) and also preserving content information (maximizing the s-BLEU between the source and style-masked sentences). We see that though Vanilla Attention is able to generally produce the lowest Accuracy\%, it does so at the expense of preserving content, reflected as lower s-BLEU compared to EA. EA, on the other hand, has the best content-preserving style masking throughout all datasets and comes as a close second in terms of Accuracy \%. Other gradient-based methods do not perform favourably in any aspect.

\begin{table*}[t]
\centering
\small
\begin{tabular}{c|cc|cc|cc|cc} 
\hline
\multicolumn{1}{l}{}                           & \multicolumn{2}{c}{\textbf{Yelp}}                         & \multicolumn{2}{c}{\textbf{IMDb}}                         & \multicolumn{2}{c}{\textbf{Amazon}}                       & \multicolumn{2}{c}{\textbf{SNLI}}      \\ 
\hline
\multicolumn{1}{c}{\textbf{Attribution Model}} & \textbf{Acc.\%} & \multicolumn{1}{c}{\textbf{s-BLEU}} & \textbf{Acc.\%} & \multicolumn{1}{c}{\textbf{s-BLEU}} & \textbf{Acc.\%} & \multicolumn{1}{c}{\textbf{s-BLEU}} & \textbf{Acc.\%} & \textbf{s-BLEU}  \\ 
\hline
Vanilla Attention (VA)                              & 73.8                & 62.41                               & \textbf{69.8}       & 62.4                                & \textbf{70}         & 57.54                               & \textbf{50.76}      & 66               \\
Explainable Attention (EA)                          & \textbf{71.3}       & \textbf{64.32}                      & 75.25               & \textbf{70}                         & 77.36               & \textbf{73.21}                      & 66.5                & \textbf{85.14}   \\
Vanilla Gradients                              & 74.2                & 38.8                                & 81.5                & 54.47                               & 74.64               & 44.19                               & 61.36               & 39               \\
Gradients * X                                  & 97.2                & 37                                  & 93                  & 50.35                               & 84.92               & 40.37                               & 70.14               & 39               \\
Integrated Gradients                           & 77.7                & 37.29                               & 81.75               & 42.42                               & 71                  & 40.77                               & 74.73               & 43               \\
No Masking                                     & 100                 & 100                                 & 100                 & 100                                 & 100                 & 100                                 & 100                 & 100              \\
\hline
\end{tabular}
\caption{Comparison of quality of Style-Masking produced using various attribution models. We found that $\lambda_{\epsilon}=0$ worked best with all gradient-based methods. For attention based methods (VA and EA), we found that $\lambda_{\epsilon}=0.15, 0.5$ worked best for \{Yelp, IMDb, Amazon\}, SNLI respectively.}
\label{tab:attri-compare}
\end{table*}

\subsection{Effect of $\lambda_\epsilon$ on Style-Masking}
\label{sec:eps-effect}
\begin{figure}[t]
\includegraphics[scale=0.33,trim=40 50 70 60, clip]{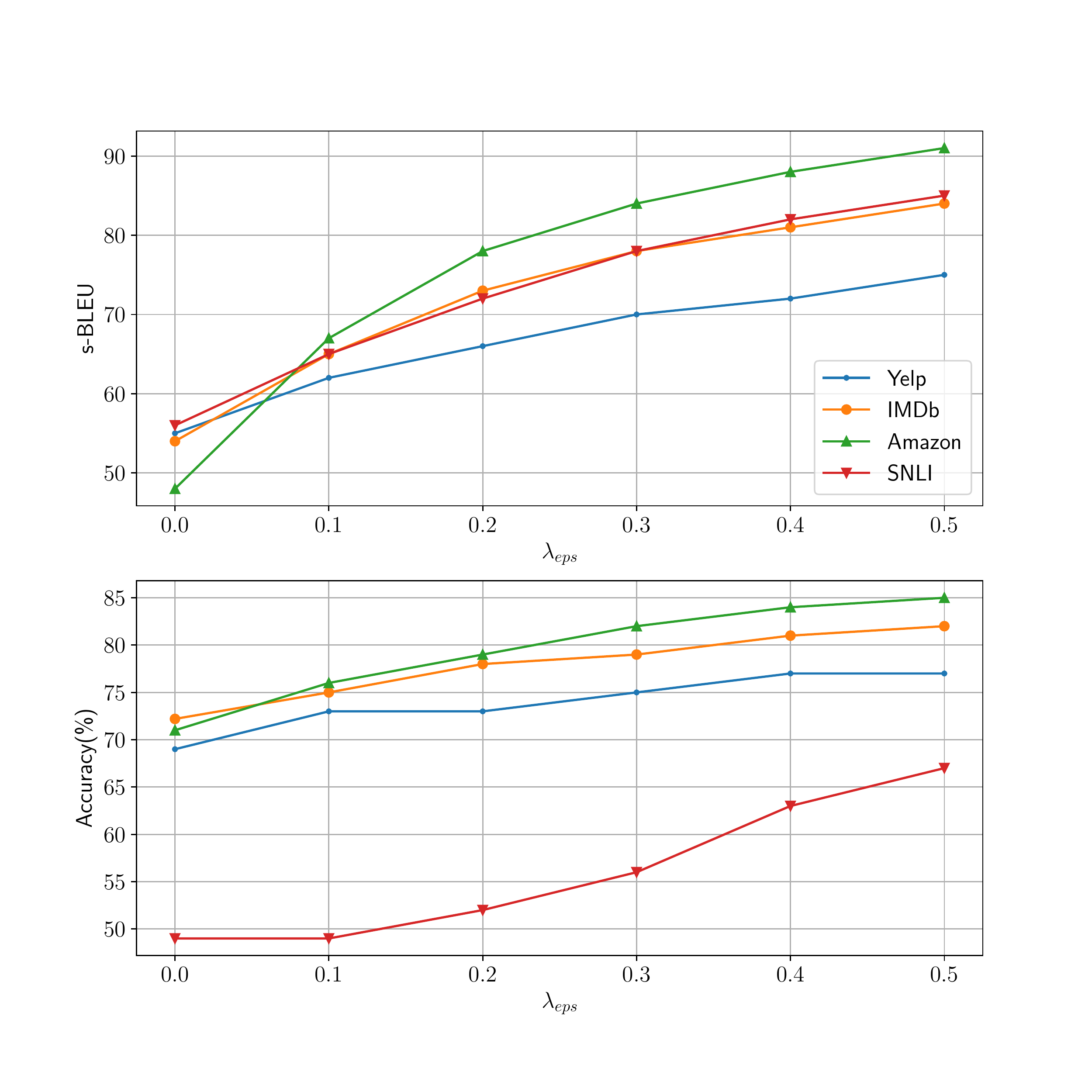}
\caption{Effect of $\lambda_\epsilon$ on the resultant style-masked sentences using our "EA+AS" method. We compute s-BLEU(Top) and Accuracy\%(Bottom) of the style-masked sentences on the test split of each dataset.}
\label{fig:eps-effect}
\vspace{-0.2in}
\end{figure}

We can intuitively reason that s-BLEU metric of the style-masked sentences serves as a rough upper bound of the s-BLEU we achieve on the output sentences after TST. 
As expected (from Eq. 1), we observe a positive correlation between $\lambda_{\epsilon}$ and both s-BLEU and Accuracy\% as seen in Table \ref{fig:eps-effect}. It is desirable to carefully choose $\lambda_{\epsilon}$ to be a high enough to boost future s-BLEU scores on the TST task and also ensure that the sentences are sufficiently style masked with low Accuracy\% scores. Only manual checking, we observed that $\lambda_{\epsilon}=0.15$ served best to accurately style-mask sentences for the Yelp, IMDb and Amazon datasets. SNLI required a higher $\lambda_{\epsilon}$ of 0.5 to ensure content information was preserved appropriately.

\section{Experiments}
We perform TST on all the datasets delineated in Section \ref{dataset_details} and analyse the results.
\subsection{Automatic Evaluation Metrics}
Typically, metrics for TST include Style Transfer Accuracy (using a pretrained classifier), BLEU for content preservation, and perplexity (using a pretrained LM) to check the fluency of outputs. \citet{cpvaee} show that this traditional set of metrics can be gamed. For fluency, we use the "Naturalness" metric \cite{naturalness} instead of PPL as it is shown to correlate better with human judgement. Apart from using BLEU for content preservation, we also report METEOR \cite{meteor}, ROUGE-L \cite{rogue} and CIDEr \cite{cider} scores in Table \ref{tab:content-metrics}.
\subsection{Baselines and Hyperparameters}
\textbf{Baseline Selection.} As baselines, we choose DirR \cite{direct}, Stable \cite{stable}, Transforming \cite{transforming}, Tag \cite{tag}, CrossAligned \cite{crossaligned}, CycleRL \cite{cycle_rl}, StyleEmbedding \cite{styleembedding}, D\&R \cite{simple} and CycleMulti \cite{cyclemulti}. For the hyperparameters of each baseline, we consider the optimal parameters of the best models for each dataset reported in each respective work. Whenever available, we directly make use of the flagship TST outputs published as part of the original work of each reference paper to ensure that a fair comparison is done.
\\
\textbf{SMLM Versions. }We consider three version of the SMLM model i.e. without fine-tuning (SMLM), with fine-tuning for 1 epoch (SMLM-F). We also consider a Generational version of SMLM with an additional auto-regressive Transformer-based decoder (SMLM-G). 

\textbf{SMLM Hyperparameters. } SMLM consists of a lightweight Transformer Encoder block with 2 layers, 8 heads and embeddings of size 512. We find that training it for 15 epochs on the Same-Style Reconstruction Task and fine-tuning it for 1 epoch on for TST produces optimal results. During fine-tuning, $\lambda_{sta}$ was set to 1 and gradient-clipping with a threshold of $10^{-3}$ was set to prevent gradients from exploding.

\textbf{Diversity-LSTM Hyperparameters. } As mentioned earlier, we use the "explainable" attention scores of the Diversity-LSTM classifier in our style-masking step. The Diversity-LSTM's mechanism, auxiliary losses and hyperparameters chosen is presented in Section \ref{sec:ea}.

\textbf{Setup and \#Parameters. } 
The Tag \cite{tag} and DirR \cite{direct} models (the two best performing baselines) have 50M and 1.5B parameters respectively. The SMLM has 45M parameters, 30x lesser parameters than DirR's fine-tuned GPT-2 model, and roughly the same number of parameters as Tag, but outperforming it in the IMDb and SNLI datasets. We report details on training time required and infrastructure used in Section \ref{sec:infra}.

\subsection{Quant. metrics for TST task}
We compute TST\% (percentage of sentences with target style) using a Bi-LSTM based pretrained classifier trained on each dataset (refer \ref{class-stats} for classifier details). r-BLEU and s-BLEU refer to the BLEU score taken between the output sentences and the human reference and ground truth sentences, respectively. For fluency, we measure the mean "Naturalness" score (the "Nat." column) as the mean classification score of a pretrained fluency discriminator \footnote{We use the pretrained naturalness classifier available in https://github.com/passeul/style-transfer-model-evaluation.}. We also add a "Mean" score consisting of the average of TST\%, s-BLEU and Nat. (normalised to 100) columns to denote a rough measure of the overall quality of each TST model. 
\subsection{Sentiment TST}
Sentiment TST is performed on Yelp (Table \ref{tab:yelp-quant}), IMDb (Table \ref{tab:imdb-quant}) and Amazon (Table \ref{tab:amazon}). We observe that for Yelp and IMDb, SMLM-FT and DirR are the best performing models according to the Mean score. \textbf{In IMDb}, DirR performs better than SMLM-FT in content preservation metrics but slightly lags behind in naturalness scores. 
\textbf{In IMDb}, SMLM-FT achieves a significantly high TST\% score at a reasonable s-BLEU of 60.9. \textbf{In Amazon, } Tag and SMLM-FT are the best performing, with Tag having a significantly high TST\% score at competitive content and fluency metrics. Overall we observe SMLM-FT, DirR and Tag as the best performing models.

\subsection{Discourse TST}
It might be presumed that Prototype editing methods such as SMLM are only capable of working well on "course-grained" styles, i.e. where the presence of style is determined by the presence of a set of certain words that convey the style.  To inspect if this is true and gauge the ability of the SMLM to operate on more cognitive and complex tasks, we consider "Discourse TST" by using Natural Language Inference (NLI) datasets.

We report statistics for the SMLM in Table \ref{tab:snli-quant}. We observe that the SMLM-FT model does well overall in this task and obtains a strong mean score of 85.53 (higher than the sentiment TST tasks). It lags behind a little mainly in the TST\% metric, but with a strong s-BLEU and Naturalness score. We also report qualitative examples of the discourse TST task in Table \ref{tab:snli-qual} and \ref{tab:more-qual}.

\begin{table}
\centering
\small
\setlength{\tabcolsep}{3pt} 

\begin{tabular}{cccccc} 
\hline
\textbf{Model} & \textbf{TST\%} & \textbf{r-BLEU} & \textbf{s-BLEU} & \textbf{Nat.} & \textbf{Mean}  \\ 
\hline
DirR           & 92.9           & 23.5            & 60.8            & 0.84                 & \textbf{79.27}          \\
Stable         & 81.6           & 15.6            & 39.2            & 0.73                 & 64.6           \\
Transforming   & 84.8           & 18.1            & 44.7            & 0.83                 & 70.9           \\
Tag            & 87.7           & 16.9            & 47              & 0.83                 & 72.57          \\
CrossAligned   & 74.4           & 6.8             & 20.2            & 0.68                 & 54.2           \\
CycleRL        & 51.1           & 14.8            & 46.1            & 0.86                 & 61.07          \\
StyleEmbedding & 8.59           & 16.7            & 67.6            & 0.87                 & 54.4           \\
D\&R     & 88             & 12.6            & 36.8            & 0.89                 & 71.27          \\
cycleMulti     & 83.8           & 22.5            & 63              & 0.86                 & \textbf{77.6}           \\ 
\hline
SMLM           & 56.5           & 20.5            & 63.2            & 0.85                 & 68.23          \\
SMLM-G         & 63.4           & 20.3            & 61.3            & 0.83                 & 69.23          \\
SMLM-FT        & 91.2           & 18.3            & 53.4            & 0.88                 & \textbf{77.6}           \\
\hline
\end{tabular}
\caption{Quantitative metrics for the Yelp Dataset.}
\label{tab:yelp-quant}
\end{table}
\begin{table}
\centering
\small
\setlength{\tabcolsep}{3pt} 
\begin{tabular}{cccccc} 
\hline
\textbf{Model} & \textbf{TST\%} & \textbf{r-BLEU} & \textbf{s-BLEU} & \textbf{Nat.} & \textbf{Mean}  \\ 
\hline
DirR           & 58.2           & 30.1            & 60.6            & 0.91                 & 69.93          \\
Stable         & 57.2           & 24.9            & 50              & 0.83                 & 63.4           \\
Transforming   & 58.7           & 25.5            & 52.3            & 0.92                 & 67.5           \\
Tag            & 75             & 32.6            & 68.3            & 0.91                 & \textbf{78.1}           \\
CrossAligned   & 73.9           & 1.5             & 2.5             & 0.62                 & 46.1           \\
StyleEmbedding & 41.1           & 13.4            & 31.2            & 0.92                 & 54.8           \\
DR             & 52             & 27.2            & 56.6            & 0.92                 & 66.9           \\ 
\hline
SMLM           & 52.4           & 31              & 71              & 0.91                 & \textbf{71.47}          \\
SMLM-G         & 53.4           & 31              & 69.6            & 0.88                 & 70.33          \\
SMLM-FT        & 63.9           & 29.6            & 69.8            & 0.92                 & \textbf{75.03}          \\
\hline
\end{tabular}
\caption{Quantitative metrics for the Amazon dataset.}
\label{tab:amazon}
\end{table}
\begin{table}
\centering
\small
\setlength{\tabcolsep}{3pt} 
\begin{tabular}{ccccc} 
\hline
\textbf{Model}      & \textbf{TST\%} & \textbf{s-BLEU} & \textbf{Nat.} & \textbf{Mean}  \\ 
\hline
DirR       & 80.3           & 67.9            & 0.92                 & \textbf{80.05}          \\
cycleMulti & 67.2           & 73.7            & 0.93                 & \textbf{77.95}          \\ 
\hline
SMLM       & 66.8           & 69.2            & 0.92                 & 76             \\
SMLM-G     & 68.9           & 65.6            & 0.93                 & 75.83          \\
SMLM-FT    & 87.9           & 60.9            & 0.92                 & \textbf{80.27}          \\
\hline
\end{tabular}
\caption{Quantitative metrics for the IMDb dataset.}
\label{tab:imdb-quant}
\vspace{-0.2in}
\end{table}
\begin{table}
\centering
\small
\setlength{\tabcolsep}{3pt} 
\begin{tabular}{ccccc} 
\hline
\textbf{Model} & \multicolumn{1}{l}{\textbf{TST\%}} & \multicolumn{1}{l}{\textbf{s-BLEU}} & \multicolumn{1}{l}{\textbf{Nat.}} & \multicolumn{1}{l}{\textbf{Mean}}  \\ 
\hline
Tag           & 48.3             & 90.2               & 0.98                    & 78.83              \\ 
\hline
SMLM           & 52.2                               & 88.5                                & 0.98                                     & 79.57                              \\
SMLM-G         & 58                                 & 86.7                                & 0.98                                     & 80.9                               \\
SMLM-FT        & 76.3                               & 86.3                                & 0.94                                     & \textbf{85.53}                              \\
\hline
\end{tabular}
\caption{Quantitative metrics for the SNLI dataset.}
\label{tab:snli-quant}
\end{table}

\subsection{Ablation Studies}
\label{sec:ablation}
\textbf{Without Finetuning. } As expected, The "SMLM" model in Table \ref{tab:yelp-quant}, \ref{tab:imdb-quant}, \ref{tab:amazon} and \ref{tab:snli-quant} has substantially better r-BLEU and s-BLEU scores compared to the fine-tuned SMLM but lacks in TST\% metrics.
\\
\textbf{Using Decoder. } The generational non fine-tuned "SMLM-G" model performs quite similarly to the non-fined-tuned decoder-less "SMLM" model with respect to the mean score. Practically, we observed that trying to fine-tune SMLM-G for a better TST\% metric is difficult and leads to gradient exploding even with a large amount of gradient clipping. The SMLM-FT model appears to be more stable during fine-tuning with a lesser computational expense.

\subsection{Additional Content Metrics.}
Past work does not tend to clarify the meaning and prioritise the presence of "content-preservation" abilities in TST models \citet{enhancingcp}.
In this effort, a more thorough analysis of content preservation abilities of DirR, Tag and SMLM-FT is given in Section \ref{tab:content-metrics}.

\subsection{Qualitative Examples}
Examples of the TST task performed using the SMLM for the IMDb and SNLI dataset are given in Table \ref{tab:imdb-qual} and \ref{tab:snli-qual} respectively.

\subsection{Human Evaluations}
We only consider the relatively unexplored Discourse TST (Entailment to Contradiction and vice versa) task for human evaluations. We were unable to reproduce the DirR baseline to run over the SNLI dataset. Therefore, we only compare the next strongest performing models i.e., Tag and SMLM. Three volunteers were given the task of voting on 200 instances (equally split for the E to C and C to E task) from the test set. A vote consists of four options i.e., "Model 1 better", "Model 2 better" or "Both Good", "Both Bad", where the models were randomised. To determine the decision an each instance, a majority considered by taking three separate votes, one from each volunteer. In the case of no majority, "No agreement" decision was taken. As seen in Table \ref{tab:human-eval}, the SMLM performs better in both tasks by a significant margin.
We report some qualitative examples taken from the evaluation in Table \ref{tab:more-qual}.
\begin{table}[t]
\small
\centering
\begin{tabular}{cccccc} 
\hline
\textbf{Direction} & \begin{tabular}[c]{@{}c@{}}\textbf{Tag }\\\textbf{better}\end{tabular} & \begin{tabular}[c]{@{}c@{}}\textbf{SMLM }\\\textbf{better}\end{tabular} & \begin{tabular}[c]{@{}c@{}}\textbf{Both}\\\textbf{~Good}\end{tabular} & \begin{tabular}[c]{@{}c@{}}\textbf{Both }\\\textbf{Bad}\end{tabular} & \begin{tabular}[c]{@{}c@{}}\textbf{No }\\\textbf{Agreement}\end{tabular}  \\ 
\hline
\textbf{E to C}    & 11                                                                     & \textbf{55}                                                             & 8                                                                     & 18                                                                   & 8                                                                         \\
\textbf{C to E}    & 8                                                                      & \textbf{44}                                                             & 2                                                                     & 35                                                                   & 11                                                                        \\
\hline
\end{tabular}
\caption{Human Evaluations done to compare Tag and SMLM on Discourse TST task on SNLI dataset. "E" and "C" denote "Entailment" and "Contradiction" respectively.}
\label{tab:human-eval}
\vspace{-0.2in}
\end{table}

\section{Conclusion}

We introduce the SMLM, a modification of the standard MLM, which we show is capable of performing TST by using a style-masked input and performing a simple same-style reconstruction task with a lightweight Transformer Encoder block. On fine-tuning the SMLM for the TST\%, it is on par with state of the art models with orders of more parameters and sophisticated architectures in the Sentiment TST task. We show that complex styles such as flow of logic/ discourse can be manipulated even with using this simple style masking assumption. We empirically show that the SMLM performs well in this Discourses Manipulation task and outperforms another strong baseline in this task, also seen through human evaluations.

\section{Limitations}
The apparent limitation with all prototype editing models, including the SMLM, is that it encourages the model to only fill in necessary style words and preserve the length and structure of the original sentence. In the case of SMLM, the word-to-word input-output mapping while training the encoder prevents the output sentence length from changing. Though it can be argued that this even works for a relatively cognitive style like discourse, in the future, there might exist styles which explicitly require the addition/deletion of words/phrases in order to successfully alter the style. Future work will therefore focus on enable variable length TST outputs, similar to the \cite{tag} approach or by incorporating a padded masked language model \cite{padded}.
\\\\


\bibliography{anthology,custom}
\bibliographystyle{acl_natbib}

\appendix

\textbf{\large{Appendix}} 

\section*{Qualitative examples of TST}
\begin{table*}
\small
\centering
\begin{tabular}{p{1cm}p{7cm}p{7cm}} 
\hline
\textbf{Direction}    & \textbf{Negative to Positive}                                     & \textbf{Positive to Negative}                                           \\ 
\hline
\textbf{Input}        & ben affleck is back to making the same boring bad acting films .  & this movie is by far one of the best urban crime dramas i 've seen .    \\
\textbf{Style Masked} & ben affleck is back to making the same <mask> <mask> acting films .   & this movie is by <mask> one of the <mask> urban crime <mask> i 've seen .     \\
\textbf{Output}       & ben affleck is back to making the same truly great acting films . & this movie is by far one of the worst urban crime garbage i 've seen .  \\
\hline
\end{tabular}
\caption{Example of Sentiment TST on the IMDb dataset.}
\label{tab:imdb-qual}
\end{table*}
\begin{table*}
\small
\centering
\begin{tabular}{p{1cm}p{7cm}p{7cm}} 
\hline
\textbf{Direction}                                                      & \textbf{Entailment to Contradiction}                                            & \textbf{Contradiction to Entailment}                                                                    \\ 
\hline
\textbf{Input}                                                          & a guy in a red jacket is snowboarding in midair . a guy is outside in the snow  & a woman is sitting outside at a table using a knife to cut into a sandwich . a woman is sitting inside  \\
\textbf{Style Masked} & a guy in a red jacket is snowboarding in midair . a guy is <mask> in the <mask>     & a woman is sitting outside at a table using a knife to cut into a sandwich . a woman <mask> <mask> <mask>     \\
\textbf{Output}                                                         & a guy in a red jacket is snowboarding in midair . a guy is swimming in the park & a woman is sitting outside at a table using a knife to cut into a sandwich . a woman is a outside       \\
\hline
\end{tabular}
\caption{Example of Discourse TST on the SNLI dataset.}
\label{tab:snli-qual}
\end{table*}

\section*{Details of Pretrained Classifier}
We use a Bi-LSTM as our choice of classifier as it performs comparably to FastText \cite{fasttext} and outperforms it in the SNLI dataset. A comparison of the two models is given in Table \ref{class-stats}.
\label{class-stats}
\begin{table}[!htbp]
\small
\centering
\begin{tabular}{ccc} 
\hline
\textbf{Dataset}     & \textbf{Model} & \textbf{Acc.\%}  \\ 
\hline
Yelp                 & FastText       & 97.6                 \\
                     & Bi-LSTM        & 97                   \\ 
\hline
IMDb                 & FastText       & 99.35                \\
                     & Bi-LSTM        & 99                   \\ 
\hline
Amazon               & FastText       & 92.1                 \\
                     & Bi-LSTM        & 93                   \\ 
\hline
SNLI                 & FastText       & 72.5                 \\
\multicolumn{1}{l}{} & Bi-LSTM        & 84                   \\
\hline
\end{tabular}
\caption{Comparison of FastText, Bi-LSTM models for classification task on all datasets.}
\label{tab:classifier-stats}
\end{table}

\section*{Additional Content Preservation Metrics}
We present more content preservation metrics in Table \ref{content-pres-metrics} to compare the top three performing models i.e., SMLM, Tag \cite{tag} and DirR \cite{direct}.
\label{content-pres-metrics}
\begin{table*}[t]
\small
\centering
\begin{tabular}{cccccccc} 
\hline
\thead{\textbf{Dataset}} & \thead{\textbf{Model}} & \thead{\textbf{METEOR}} & \thead{\textbf{ROUGE-L}} & \thead{\textbf{CIDEr}} & \thead{\textbf{Embedding Avg.} \\ \textbf{Cosine Sim.}} & \thead{\textbf{Vector Extrema} \\ \textbf{Cosine Sim.}} & \thead{\textbf{Greedy} \\ \textbf{Matching Score}} \\

\hline
Yelp                 & SMLM-FT           & 0.376           & 0.739            & 4.934          & 0.939                               & 0.767                               & 0.867                           \\
                     & DirR           & 0.444           & 0.83             & 5.813          & 0.969                               & 0.867                               & 0.926                           \\
\multicolumn{1}{l}{} & Tag            & 0.362           & 0.707            & 4.326          & 0.934                               & 0.765                               & 0.867                           \\
IMDb                 & SMLM-FT           & 0.414           & 0.8              & 5.657          & 0.96                                & 0.755                               & 0.891                           \\
                     & DirR           & 0.472           & 0.852            & 6.344          & 0.978                               & 0.847                               & 0.933                           \\
Amazon               & SMLM-FT           & 0.464           & 0.868            & 6.725          & 0.964                               & 0.782                               & 0.921                           \\
\multicolumn{1}{l}{} & DirR           & 0.469           & 0.823            & 6.612          & 0.967                               & 0.821                               & 0.929                           \\
                     & Tag            & 0.453           & 0.835            & 6.548          & 0.966                               & 0.781                               & 0.917                           \\
SNLI               & SMLM-FT           & 0.563           & 0.906            & 8.297          & 0.986                               & 0.886                               & 0.96                           \\
\multicolumn{1}{l}{} & Tag           & 0.606           & 0.944            & 8.619          & 0.992                               & 0.921                               & 0.972                           \\

\hline
\end{tabular}
\caption{Content Preservation metrics for all datasets comparing top performing models}
\label{tab:content-metrics}
\end{table*}

\section*{Additional examples of Discourse TST}
We focus more again on the Discourse TST task and report more examples of the SMLM and the Tag \cite{tag} baseline in Table \ref{tab:more-qual}.
\label{more-qual}
\begin{table*}
\small
\centering
\begin{tabular}{cp{6cm}p{6cm}} 
\hline
\textbf{Direction}                   & \textbf{Entailment to Contradiction}                                                                                  & \textbf{Contradiction to Entailment}                                                               \\ 
\hline
\textbf{Input}                       & \textbf{a black women holding a sign that says free hugs in the city . a woman is holding a sign} & \textbf{a tan dog chases a black and white soccer ball . a dog is chasing after a cat}             \\
Output (Tag)                         & a black women holding a sign that says free hugs in the city . a woman is holding a sign      & a tan dog chases a black and white soccer ball . a dog is chasing after a sport                    \\
Output (SMLM-FT)                     & a black women holding a sign that says free hugs in the city . a woman is holding a book          & a tan dog chases a black and white soccer ball . a dog is outside after a ball                     \\ 
\hline
\textbf{Input}                       & \textbf{a man is doing a task by a body of water on a farm . the man is doing something by the water}                 & \textbf{a dad with his child and an apple pie . a dad and his daughter with an blueberry pie}      \\
Output (Tag)                         & a man is doing a nap by a body of water on a farm . the man is doing pushups by the water                             & a dad with his child and an apple outside . a dad and his daughter with an acousticelling outside  \\
\multicolumn{1}{l}{Output (SMLM-FT)} & a man is doing a task by a body of water on a farm . the man is doing nothing by the beach                            & a dad with his child and an apple pie . a dad and his daughter with an apple outside               \\
\hline
\end{tabular}
\caption{Examples of Discourse TST on SNLI of SMLM vs Tag \cite{tag}}
\label{tab:more-qual}
\end{table*}

\section*{Computational Expense and Infrastructure used}
\label{sec:infra}
The most parameter-heavy SMLM model model was from the SNLI dataset. Therefore we report statistics for this model to gauge the overall computational expenses the SMLM demands.
The model has 45 million parameters and each epoch took approximately 224 seconds to train on an Nvidia V100-SMX2 GPU and an Intel(R) Xeon(R) E5-2698 CPU. For complete details, we will make the code open source which will also contain the models we trained along with log files with all metadata about the model architecture and training.

\section*{Ethics Statement}
\label{sec:ethics}
Any TST model can be used for illicit purposes. Therefore, it is important we keep in mind a code of ethics (e.g. \url{https://www.acm.org/code-of-ethics}). We will make all our code open-source and will contain all details of experimentation and implementation, training time, additional hyperparameters used in the form of log files included inside the directories of our saved models, which can also be used to replicate results.

\section*{The Diversity-LSTM and Explainable Attention}
\label{sec:ea}

Effective style-masking requires an attribution model with a high degree of plausibility, which motivates our use of "explainable" attention scores \citet{mohankumar} as choice for the style-masking step.
\\
\textbf{Why not use standard attention?}
Vanilla attention scores do not serve as accurate attribution scores. Attention scores over RNN hidden states for the classification task do not correlate well with other standard interpretation metrics \cite{jain}, such as gradient and occlusion based methods. 
Feeding alternative adversarial/random attention distributions lead to only a modest effects are the model's decision \cite{sarah}. 
However \citet{sarah} shows that these adversarial distributions, if properly produced, do induce poorer performance showing that vanilla attention is still partially faithful to its explanation. 
\citet{mohankumar} postulate that attention scores over hidden states (H) are not explainable due to information mixing and subsequent entanglement/coupling and mutual information among H in RNNs. To mitigate this entanglement, diversity driven learning (inspired by results in \citet{nema2}) is enforced among H. 
This promotes the attention mechanism over such diversity-enforced H to satisfy "faithfulness" and "plausibility" properties when interpreted as attribution scores, which we refer to as "Explainable attention" (EA).  \citet{mohankumar} empirically show that EA does not suffer any loss in performance in the downstream task. Supporting plausibility, a) EA scores correlate better with strong attribution tools such as Integrated Gradients b) On analysis over POS tags, EA attends more to tags which are contextually important w.r.t the given task and c) Correlates better to human judgement than vanilla attention. 

\textbf{The Diversity Driven LSTM. }
The Diversity LSTM consists of an LSTM-based classifier with attention \cite{bahdanau} over the H. The final context vector is fed through a feedforward layer to generate the output. 
\begin{align*}
    \nonumber \tilde{\alpha}_t  &= \mathbf{v}^T \mbox{tanh}(\mathbf{W}\mathbf{h} + \mathbf{b}) \; \;  \forall t \in [1,m]\\
    \nonumber \alpha_t &= \mbox{softmax}(\tilde{\alpha}_t) \\
    \nonumber \mathbf{c}_{\alpha} &= \sum_{t=1}^m \alpha_t {\mathbf{h}}_t
\end{align*}
To enforce the H of the LSTM to be "diverse" i.e. more disentangled w.r.t each other, the conicity (\citet{chandrahas}, \citet{sai}) metric is used as an auxillary loss and is defined as the mean of "Alignment to Mean" (ATM) for all vectors $\mathbf{v}_i \in \mathbf{V}$:
\begin{align}
  \nonumber \text{ATM}(\mathbf{v}_i, \mathbf{V}) = \text{cosine}(\mathbf{v}_i, \frac{1}{m} \sum_{j=1}^{m} \mathbf{v}_j)
  \\
  \nonumber \text{conicity}(\mathbf{V}) =  \frac{1}{m} \sum_{i=1}^{m} \text{ATM}(\mathbf{v}_i, \mathbf{V})
\end{align}
The attention mechanism over a Diversity LSTM's H is now encouraged to be faithful to a particular set of scores, thus promoting the model to move towards more faithful and plausible attributions.
The final loss is given as:
\begin{align}
    \nonumber L(\theta_{Div}) =  - \mbox{log} p_{Div} (y | P) +  \lambda_con \; \mbox{conicity}(\mathbf{H}^{P})
\end{align}

EA requires only training an additional diversity driven RNN classifier over the given dataset. After which, a single forward pass is required to obtain attribution scores. This is unlike other methods such as IG, Lime, DeepLift, Occlusion, wherein each generating each explanation requires comparatively more operations.

\section*{Diversity-LSTM Hyperparameter Selection}
Table \ref{div-lstm-hyp} gives details of the Diversity-LSTM classifier used during the style-masking step. We also note that the Diversity-LSTM's performance is comparable to a Bi-LSTM and FastText classifier (as shown in \ref{class-stats}).
\label{div-lstm-hyp}
\begin{table}[!htbp]
\small
\centering
\begin{tabular}{cccc} 
\hline
\textbf{Dataset} & \textbf{Acc.\%} & \textbf{$Loss_{con}$} & \textbf{$\lambda_{con}$}  \\ 
\hline
Yelp             & 96                  & 0.06                   & 10                 \\
IMDb             & 100                 & 0.09                   & 10                 \\
Amazon           & 89                  & 0.03                   & 20                 \\
SNLI             & 82                  & 0.18                   & 10                 \\
\hline
\end{tabular}
\caption{Classification task statistics and choice of $\lambda_{con}$ for each dataset.}
\label{tab:div-lsmt-stats}
\end{table}

\FloatBarrier

\begin{table}[!htbp]
\small
\centering
\begin{tabular}{ccccc} 
\toprule
Dataset & Style Label                                                  & Train                                                & Dev                                                  & Test                                                  \\ 
\hline
Yelp    & \begin{tabular}[c]{@{}l@{}}Positive \\Negative\end{tabular} & \begin{tabular}[c]{@{}l@{}}266K \\177K\end{tabular} & \begin{tabular}[c]{@{}l@{}}2000 \\2000\end{tabular} & \begin{tabular}[c]{@{}l@{}}500 \\500\end{tabular}    \\ 
\hline
IMBb    & \begin{tabular}[c]{@{}l@{}}Positive \\Negative\end{tabular} & \begin{tabular}[c]{@{}l@{}}178K \\187K\end{tabular} & \begin{tabular}[c]{@{}l@{}}2000 \\2000\end{tabular} & \begin{tabular}[c]{@{}l@{}}1000 \\1000\end{tabular}  \\ 
\hline
Amazon  & \begin{tabular}[c]{@{}l@{}}Positive\\Negative\end{tabular} & \begin{tabular}[c]{@{}l@{}}277K \\179K\end{tabular} & \begin{tabular}[c]{@{}l@{}}985 \\1015\end{tabular}  & \begin{tabular}[c]{@{}l@{}}500 \\500\end{tabular}    \\
\hline
SNLI    & \begin{tabular}[c]{@{}l@{}}Entailment\\ Contradiction\\\end{tabular} & \begin{tabular}[c]{@{}l@{}}183K \\183K\end{tabular} & \begin{tabular}[c]{@{}l@{}}3329 \\3278\end{tabular} & \begin{tabular}[c]{@{}l@{}}3368 \\3237\end{tabular}  \\ 

\bottomrule
\end{tabular}
\caption{Split and label wise statistics of each dataset.}
\label{dataset_details}
\end{table}
\FloatBarrier

\section*{Training Algorithm}
\label{sec:algo}
\begin{algorithm}[!htbp]
	\ForEach{bootstrap\_epoch}{
		\ForEach{mini-batch}{
		minimize $L_{BS}$ w.r.t $\theta_{SMLM}$
	
	    }
	}
	
	\ForEach{finetune\_epoch}{
		\ForEach{mini-batch}{
		minimize $L_{BS}$ w.r.t $\theta_{SMLM},\theta_{cls}$ \\
		minimize $L_{cls}$ w.r.t $\theta_{cls}$
	
	    }
	}
	\caption{Training process.}
	\label{alg:training-process}
\end{algorithm}

\end{document}